\documentclass[conference]{IEEEtran}
\IEEEoverridecommandlockouts
\usepackage{cite}
\usepackage{svg}
\usepackage{comment}
\usepackage{subfig}
\usepackage{amsmath,amssymb,amsfonts}
\usepackage{algorithmic}
\usepackage{graphicx}
\usepackage{textcomp}
\usepackage{tikz}
\usetikzlibrary{shapes.geometric, arrows}

\definecolor{OrangeRed}{RGB}{255,69,0}
\definecolor{ProcessBlue}{RGB}{0,176,240}
\definecolor{BurntOrange}{RGB}{204,85,0}
\definecolor{LimeGreen}{RGB}{50,205,50}

\tikzstyle{startstop} = [rectangle, rounded corners,
minimum height = 1cm,
text centered, draw=black, fill=OrangeRed!50]
\tikzstyle{io} = [trapezium, trapezium stretches=true,
trapezium left angle=70, trapezium right angle=110,
minimum height = 1cm,
text centered, draw=black, fill=ProcessBlue!50]
\tikzstyle{process} = [rectangle, 
minimum height = 1cm,
text centered, draw=black, fill=BurntOrange!50]
\tikzstyle{decision} = [diamond, aspect=2, 
minimum height = 1cm,
text centered, draw=black, fill=LimeGreen!50]
\tikzstyle{arrow} = [thick,->,>=stealth]
\def\BibTeX{{\rm B\kern-.05em{\sc i\kern-.025em b}\kern-.08em
    T\kern-.1667em\lower.7ex\hbox{E}\kern-.125emX}}

\begin{document}
\title{From Natural Language to Solver-Ready Power System Optimization: An LLM-Assisted, Validation-in-the-Loop Framework}
\author{
\begin{minipage}[t]{0.32\textwidth}
\centering
\text{Yunkai Hu}\\
\textit{Department of Electrical and Computer Engineering}\\
\textit{University of Southern California}\\
Los Angeles, CA \\
yunkaihu@usc.edu
\end{minipage}
\hfill
\begin{minipage}[t]{0.32\textwidth}
\centering
\text{Tianqiao Zhao}\\
\textit{Department of Electrical Engineering}\\
\textit{University of Texas at Arlington}\\
Arlington, TX \\
tianqiao.zhao@uta.edu
\end{minipage}
\hfill
\begin{minipage}[t]{0.32\textwidth}
\centering
\text{Meng Yue}\\
\textit{Interdisciplinary Science Department}\\
\textit{Brookhaven National Laboratory}\\
Upton, NY \\
yuemeng@bnl.gov
\end{minipage}
}

\maketitle

\begin{abstract}
This paper introduces a novel Large Language Models (LLMs)-assisted agent that automatically converts natural-language descriptions of power system optimization scenarios into compact, solver-ready formulations and generates corresponding solutions. In contrast to approaches that rely solely on LLM to produce solutions directly, the proposed method focuses on discovering a mathematically compatible formulation that can be efficiently solved by off-the-shelf optimization solvers. Directly using LLMs to produce solutions often leads to infeasible or suboptimal results, as these models lack the numerical precision and constraint-handling capabilities of established optimization solvers. The pipeline integrates a domain-aware prompt and schema with an LLM, enforces feasibility through systematic validation and iterative repair, and returns both solver-ready models and user-facing results. Using the unit commitment problem as a representative case study, the agent produces optimal or near-optimal schedules along with the associated objective costs. Results demonstrate that coupling the solver with task-specific validation significantly enhances solution reliability. This work shows that combining AI with established optimization frameworks bridges high-level problem descriptions and executable mathematical models, enabling more efficient decision-making in energy systems.
\end{abstract}


\section{Introduction}
Classical optimization remains the backbone of power-system decision making. Unit commitment (UC), economic dispatch, and optimal power flow are routinely expressed as mixed-integer and nonlinear programs and solved with mature engines that provide feasibility enforcement, dual information, and—where applicable—optimality certificates \cite{wood2013pgoc}. Decades of work on decomposition, cutting planes, Lagrangian relaxations, and stochastic/robust variants have pushed these formulations to industrial scale while preserving auditability and model transparency \cite{yang2021machine}. The principal drawback is the front end: translating evolving operational policies into mathematically precise, solver-compatible models is labor-intensive and error-prone, which slows iteration and limits accessibility for non-experts.

AI/ML approaches have sought to reduce this burden or accelerate solve times \cite{perera2019machine}. Supervised surrogates approximate AC-OPF mappings or screen active constraint sets\cite{guha2019mlacopf,pan2020deepopf}; reinforcement learning (RL) has been explored for scheduling and corrective actions \cite{10246411}; and “learning to optimize” techniques provide warm starts, learned cuts, or heuristics embedded in the solver loop \cite{qin2025solve,yang2025stable}. These methods can deliver speedups or good average-case performance within the training distribution, but they often trade certainty for speed: constraint violations can slip through, guarantees are scarce, and behavior may degrade under distribution shift. Even when constraints are enforced via projection or penalty terms, certification and interpretability remain challenging, which limits adoption in mission-critical operations.

More recently, Large Language Models (LLMs) have been used to “solve” grid operations problems directly from text by reasoning step-by-step and outputting candidate schedules or dispatch vectors \cite{cheng2025large,huang2024large,zhou2024elecbench,zhang2025review}. While appealingly simple, this direct mode collides with the numerical precision and strict feasibility demands of power-system optimization. Directly using LLMs to produce solutions often leads to infeasible or suboptimal results, as these models lack the numerical precision and constraint-handling capabilities of established optimization solvers. As problem scale grows and constraints tighten—e.g., minimum up/down times, ramping, spinning reserves, and network limits—the outputs become inconsistent and non-reproducible, and there is no principled way to certify optimality or even feasibility.

Our work takes a different path by using the LLM not as a solver but as a \emph{formulation generator} \cite{ahmaditeshnizi2024optimus}: an agent built on LLMs that converts natural-language scenario descriptions into compact, solver-ready mathematical programs and then delegates computation to off-the-shelf solvers. The pipeline is built around domain-aware prompting and a structured schema so that the LLM emits typed variables, objective terms, and constraints with explicit data bindings rather than free-form text. A validation-and-repair layer detects omissions and inconsistencies—such as missing reserve coupling, incorrect ramp logic, or commitment integrality—and engages the LLM to patch the model before any optimization is attempted. By deferring numerical work to mature solvers, the agent preserves feasibility and leverages optimality certificates while still delivering the usability of natural-language interaction. In addition, we contribute a \emph{solution-enhancement} layer that accelerates branch-and-cut without altering problem semantics: a novel GNN-based policy proposes branching priorities for commitment binaries \cite{gasse2019exact}, and a lightweight LLM configures cut separators per instance within a guarded, solver-parameter interface \cite{lawless2025llms}. Both mechanisms are optional, preserve correctness, and complement the translation layer—together yielding reliable schedules and costs faster, in a manner that is auditable, extensible, and readily generalizable beyond UC.

\section{Problem Formulation}
We pose the task as translating a natural-language scenario \(S\) into a solver-ready mixed-integer program \(\mathcal{M}(S)=\{\text{sets},\text{parameters},\text{variables},\text{objective},\text{constraints}\}\). From \(S\), our agent extracts entities (e.g., generators \(I\), horizon \(T\)), attributes (capacities, costs, ramps, minimum up/down times), system requirements (demand, reserves), and policies (must-run, mutual exclusivity). An LLM produces a typed schema that binds these elements to a canonical template; a validator enforces completeness and logical consistency and, if needed, triggers iterative repair before passing \(\mathcal{M}(S)\) to a MILP solver.

Taking unit commitment (UC) as an example, the model determines each unit’s on/off status and dispatch over \(T\) to meet demand at minimum cost \cite{carrion2006milpuc}. UC combines binary commitment variables with continuous dispatch, accounting for variable, start-up, and no-load costs. Constraints include power balance with reserves, capacity limits, minimum up/down times, ramp limits, network deliverability, fuel or emissions limits, and coupling to renewables or storage. Although problem size grows with \(|I| \times |T|\), modern MILP solvers handle moderate instances efficiently with robust encodings.

ML/AI accelerators \cite{bengio2021ml4co}—such as surrogates, RL schedulers, and learning-to-branch/cut—can reduce runtime but often lack feasibility or optimality guarantees, degrade under distribution shift, and require case-specific tuning \cite{gasse2019exact}. LLM-based code generation improves accessibility but is brittle for UC: prompts often miss or mis-specify constraints \cite{chen2021evaluating}, while direct LLM solution generation is non-deterministic, scales poorly, and offers no guarantees. We address these issues with a domain-specialized, LLM-assisted translator that maps \(S\) to a validated, solver-compatible UC MILP with schema checks and iterative repair, yielding optimal or near-optimal schedules—or precise diagnostics when inputs conflict—while preserving solver rigor.

\section{LLM-Assisted Optimization Agent Overview}
\begin{figure}[htbp]
    \centerline{\includegraphics[width=7cm]{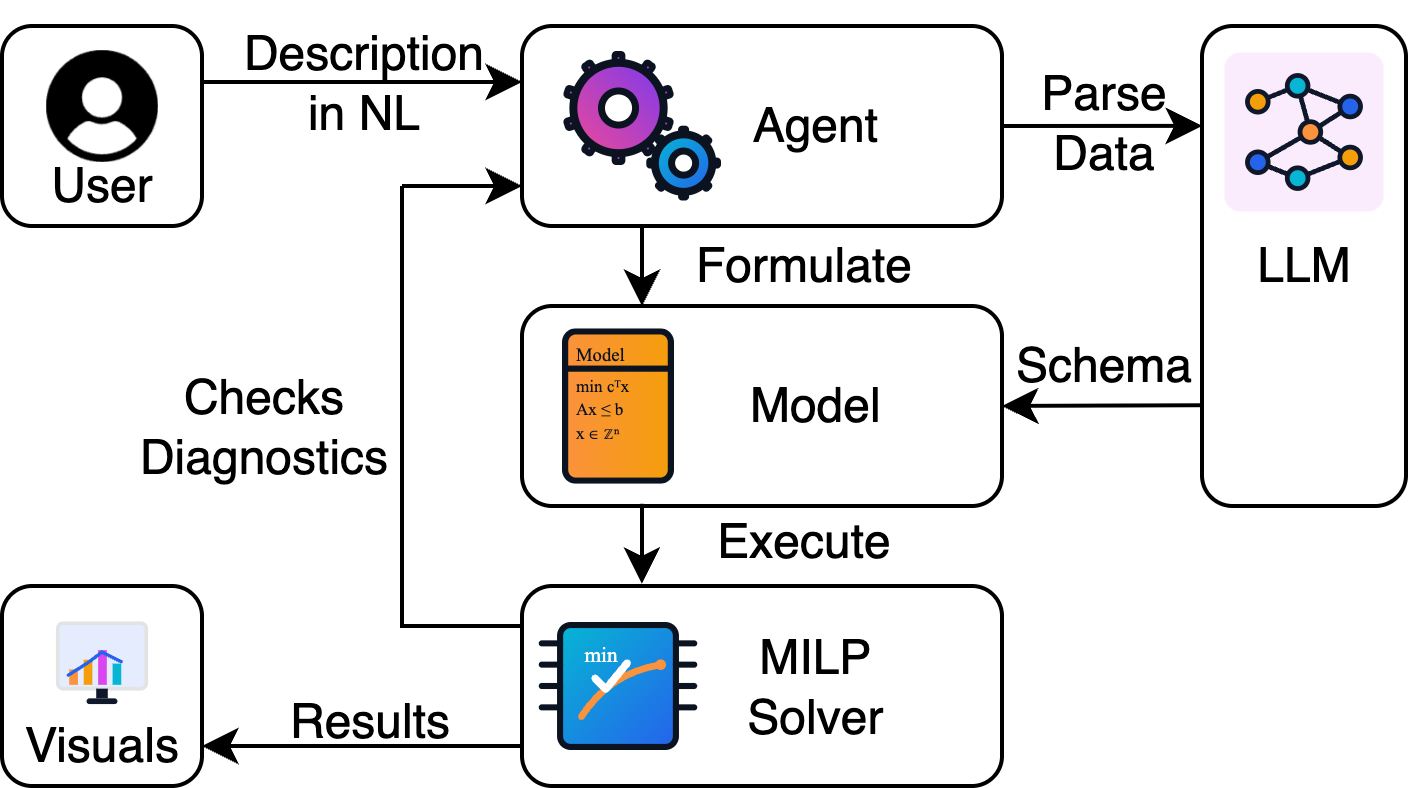}}
    \caption{High-level Framework of LLM-assisted Agent.}
    \label{fig1}
\end{figure}

The agent transforms a natural-language specification into a solver-ready model and a validated solution. As illustrated in Fig.~\ref{fig1}, a utility operator submits a scenario; an LLM-driven parser performs schema-aware parameter synthesis (entities, attributes, policies) to produce a typed model skeleton; the agent instantiates a canonical MILP formulation and hands it to the solver for execution; and the resulting schedules, costs, and diagnostics are returned through user-facing visualizations. Throughout, a validation layer enforces unit/type/shape checks and tests feasibility against the specification; on any error or violation, the agent enters a diagnosis–and–repair loop that provides targeted feedback to the LLM, amends parameters or constraints, and re-solves until success or a small iteration budget is reached.

\begin{figure}[htbp]
\centerline{
    \begin{tikzpicture}[thick,scale=0.6, every node/.style={scale=0.6}][node distance=1.8cm]
    \node (user) [startstop] {Domain expert};
    \node (nl)   [io, below of=user, yshift=-0.5cm, text width=3cm] {Problem description (natural language)};
    \node (param)[process, below of=nl, yshift=-0.5cm, text width=3cm] {Parameter synthesis\\(LLM parsing)};
    \node (chk)  [decision, below of=param, yshift=-1cm, text width=2.3cm] {Schema \& Size Checks OK?};
    \node (repair)[process, right of=chk, xshift=3.8cm, text width = 3cm] {Iterative repair\\ (Feedback to LLM)};
    \node (form) [process, below of=chk, yshift=-1cm, text width = 4cm] {Formulation Generation\\(LLM $\rightarrow$ MILP model)};
    \node (guide)[decision, below of=form, yshift=-1cm, text width = 2cm] {Guidance Enabled?};
    \node (def)  [process, right of=guide, xshift=3.5cm, text width = 3cm] {Default Solver Configuration};
    \node (gnn)  [process, left  of=guide, xshift=-3.5cm, text width = 3cm] {Branching Priorities (GNN-guided)};
    \node (solve)[process, below of=guide, yshift=-1cm, text width = 3cm] {MILP Solver (Branch-and-Cut)};
    \node (succ)[decision, below of=solve, yshift=-0.5cm] {Successful?};
    \node (sched)[io, left of=succ, xshift=-3.5cm, text width = 4cm] {Optimized Schedule \& Metrics};
    \node (debug)[decision, below of=succ, yshift=-1.2cm, text width = 2.5cm]{What Issue? (LLM Prompted)};
    \node (check_code)[process, right of=debug, xshift=3.5cm]{Check Code};
    \node (check_data)[process, below of=debug, yshift=-1cm,]{Check Data};
    \node (viz)  [startstop, below of=sched, yshift=-1cm] {Reports \& Visualization};

    \draw [arrow] (user) -- (nl);
    \draw [arrow] (nl) -- (param);
    \draw [arrow] (param) -- (chk);
    \draw [arrow] (chk) -- node[anchor=east] {yes} (form);
    \draw [arrow] (chk) -- node[anchor=south] {no} (repair);
    \draw [arrow] (repair) |- (param);
    \draw [arrow] (form) -- (guide);
    \draw [arrow] (guide) -- node[anchor=south] {yes} (gnn);
    \draw [arrow] (guide) -- node[anchor=south] {no} (def);
    \draw [arrow] (gnn) |- (solve);
    \draw [arrow] (def) |- (solve);
    \draw [arrow] (solve) -- (succ);
    \draw [arrow] (succ) -- node[anchor=south] {yes} (sched);
    \draw [arrow] (succ) -- node[anchor=east] {no} (debug);
    \draw [arrow] (debug) -- node[anchor=south]{code} (check_code);
    \draw [arrow] (debug) -- node[anchor=east]{data} (check_data);
    \draw [arrow] (check_code) |- (solve);
    \draw [arrow] (check_data.east) -- ([xshift=5cm]check_data.east) |- (solve.east);
    \draw [arrow] (sched) -- (viz);
    \end{tikzpicture}
    }
\caption{End-to-end Pipeline.}
\label{fig2}
\end{figure}
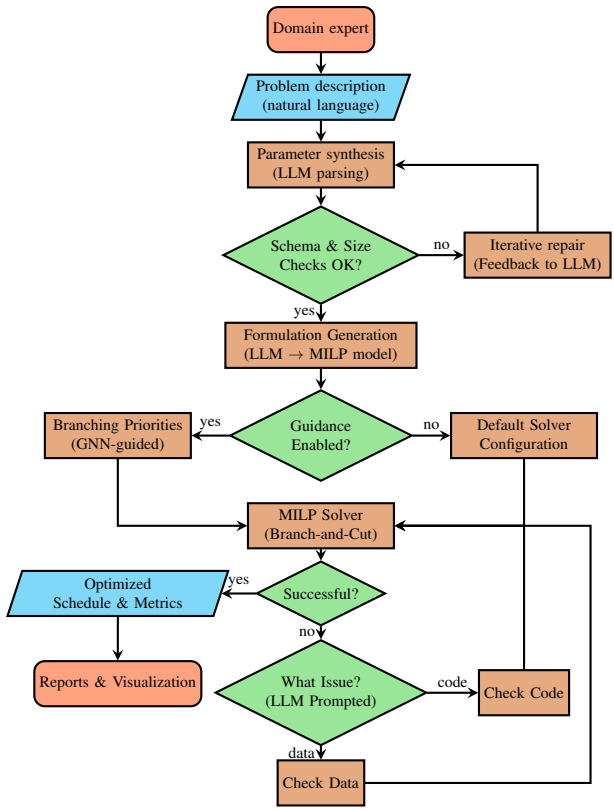

Fig.~\ref{fig2} outlines the end-to-end architecture: a domain expert supplies a scenario in plain text; an LLM-driven parser synthesizes a typed parameter schema; automatic checks verify sizes, units, and completeness, triggering an iterative repair loop when needed; a formulation module maps the schema to a solver-ready MILP; an optional guidance stage configures solver tactics; a commercial solver computes a solution; and a diagnostics loop—invoked only on failure or detected violations—pinpoints whether issues arise from code or data, repairs them, and re-solves. Successful runs produce optimized schedules and metrics, which are rendered as human-readable reports and visualizations.

\subsection{Natural Language Processing and Parameter Synthesis}
The workflow begins with a free-form scenario supplied by the user—often semi-structured as a generator table followed by operational notes and a demand profile. A light preprocessing pass standardizes units and resolves obvious inconsistencies, after which an LLM-driven parser performs schema-aware extraction of entities (generators and time horizon), attributes (capacities, costs, ramps, minimum up/down times), system requirements (load and reserves), and policy statements (must-run windows, mutual exclusivity). Rather than emitting ad hoc text, the LLM is instructed to return a typed parameter object with explicit cardinalities and array shapes. A validator then enforces type, range, and shape invariants and checks basic consistency (e.g., \(P_i^{\min}\le P_i^{\max}\), feasible ramp rates). If a check fails, targeted diagnostics are fed back to the LLM and the parameters are repaired before proceeding. By the end of this stage, the system holds the original description and a unit-harmonized, horizon-aligned schema ready for modeling.

\subsection{MILP Model Formulation}
Given the validated schema, the agent compiles a solver-ready model by prompting the LLM with a canonical UC template, the extracted data, and a few illustrative exemplars. The LLM first produces a compact, pseudo-mathematical specification that maps typed fields to variables, objective terms, and constraint families, and then renders executable code. In our implementation, code generation targets the Gurobi Python API to enable immediate execution and rapid error surfacing. Domain guidance is embedded directly in the prompt—such as enforcing capacity coupling \(p \le P_{\max}u\), respecting minimum up/down time encodings, and applying ramp trajectories at start-up/shut-down—so that common omissions are avoided without sacrificing generality.

\subsection{Optimization and Solution Retrieval}
The generated model is passed to the MILP solver together with the synthesized parameters. If the solver returns an optimal or feasible solution, the agent validates the schedules against the user’s specification to catch any residual mismatches, then assembles a concise report with per-unit commitment and dispatch, start-up/shut-down events, total cost, and reserve performance for delivery via a web dashboard. If the solver reports infeasibility or an error, or if post-solve validation flags a violation, the agent enters a diagnosis–and–repair loop: structured solver feedback (e.g., infeasible reserve hours, contradictory exclusivity rules, ramp violations around transitions) is summarized for the LLM, which proposes targeted edits to parameters or constraints; the instance is recompiled and re-solved. This loop is capped at a small iteration budget (five in our experiments) to prevent drift while providing enough room for corrective refinement.

\section{Solution Enhancement}
To reduce time-to-solution without altering problem semantics, the agent can augment the baseline branch-and-cut process with learned guidance that operates strictly through solver interfaces. Because the mathematical model is unchanged, feasibility and optimality guarantees remain intact. Two complementary options are supported: a graph neural network (GNN) that proposes instance-specific branching priorities for commitment binaries, and a lightweight LLM policy that configures cut separators prior to the solve. Both are optional and can be toggled per run.

\subsection{GNN-Based Branching Policy}
Branching choices largely determine search-tree growth. Building on the learning-to-branch paradigm~\cite{gasse2019exact}, we represent the root-node MILP as a bipartite graph over variables and constraints with features drawn from the LP relaxation and problem structure, and use a message-passing GNN to score binary variables. The model is trained offline on representative UC instances with rank-based objectives derived from strong-branching proxies or relative node-count reductions on calibration trees. At inference, scores are mapped to integer priorities (e.g., Gurobi \texttt{BranchPriority}), biasing the search toward UC-salient structure while preserving completeness and optimality. Empirically, this reduces explored nodes and wall-clock time on UC benchmarks.

\subsection{LLM-Based Separator Configuration}
Cut management is the second major lever in branch-and-cut. Following recent LLM-guided approaches~\cite{lawless2025llms}, the agent summarizes inexpensive pre-solve diagnostics—variable/constraint counts, sparsity and row/column statistics, proportions of constraint families, and root-LP gap—and requests a configuration over a constrained, whitelisted action space (enabling standard separators and setting pass counts/aggressiveness). A guard layer validates types and ranges and reverts to conservative defaults on any anomaly. The configuration is applied once, with an optional single re-evaluation if early progress stalls. This policy complements learned branching and is particularly helpful when the initial relaxation is weak.

\subsection{Prompt Template}
\begin{figure}
\centering
\subfloat[]{\label{fig3:a}\includegraphics[width=0.9\linewidth]{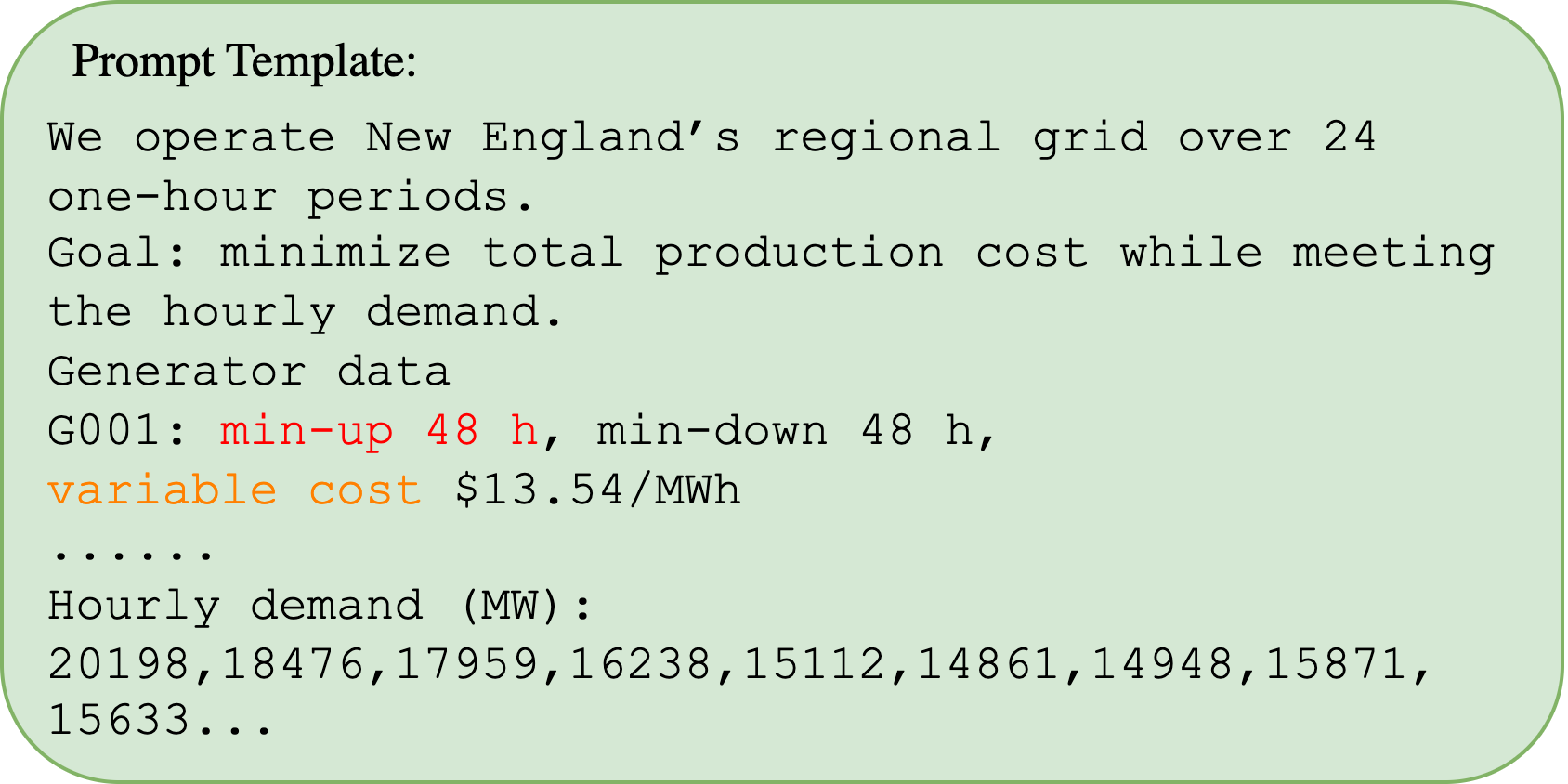}}%
\par\smallskip 
\subfloat[]{\label{fig3:b}\includegraphics[width=0.9\linewidth]{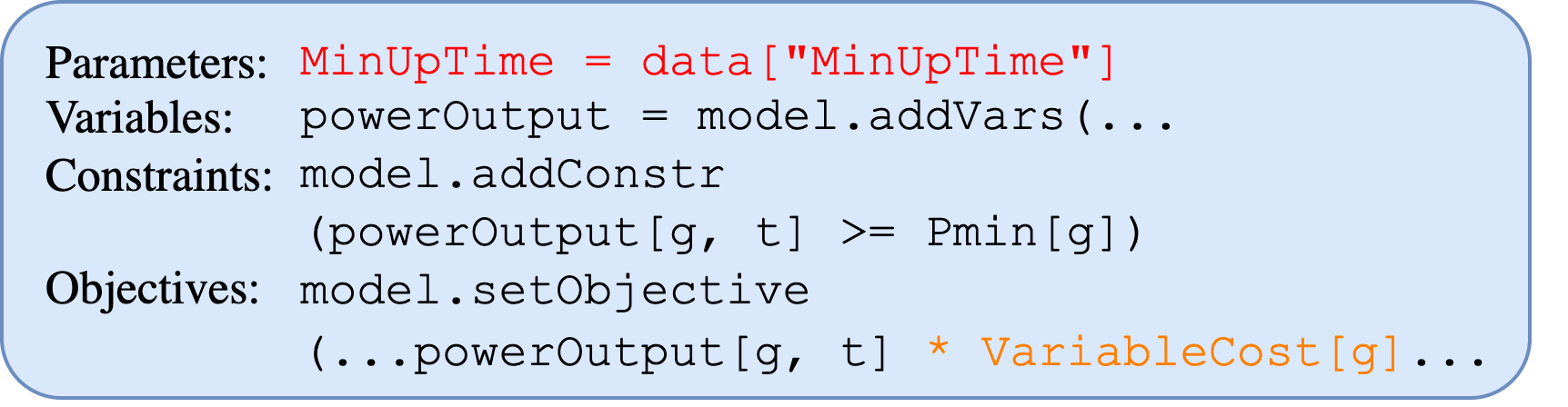}}%
\caption{Prompt Template.}
\label{fig3}
\end{figure}

Figure~\ref{fig3} illustrates the lightly structured prompt used for compilation. The template names typed fields for parameters, declares the variables to be created, and invokes canonical constraint and objective fragments, allowing the LLM to assemble a complete model deterministically from the schema. In effect, the prompt serves as a contract between natural-language intent and executable artifacts: parameters remain immutable data, variables are decision quantities, constraints encode feasibility and logic, and the objective aggregates cost terms. This structure both streamlines code generation and simplifies downstream validation.

\section{Case Study}
In this study, we take GPT-4 as the backbone agent API \cite{achiam2023gpt} and procedurally generate UC instances in a MILP-Evolve style \cite{li2024towards}, treating UC as a parametric family rather than hand-crafted cases. Each instance is a thermal UC MILP with 50--100 units partitioned into base-load, mid-merit, and peakers to induce heterogeneous costs, ramps, and start-up penalties. Two horizons are considered: a 7-day hourly horizon with weekday/weekend structure, and a 1-month daily horizon with demand and operations aggregated to daily resolution. Demand profiles combine smooth seasonal/diurnal components with stochastic perturbations and are scaled to target reserve margins. In additional, we generate a small-scale benchmark set consisting of much smaller UC instances, with only 3, 10, 30, or 60 generator over 24-hour (one day) hourly horizon. These smaller cases help evaluate our approach's performance across varying problem scales. All datasets are split by disjoint random seeds into a runtime cohort (to compare GNN-guided versus default solver speed) and an accuracy cohort (to compare objective values between expert formulations and our system). 

\subsection{Performance Metrics and Evaluation}
Each instance in the runtime cohort is solved twice—once with default solver settings and once with our GNN-guided branching (optionally with variable fixation)—and wall-clock time is recorded. We report mean and median runtime, empirical runtime distributions, and the share of instances on which guidance is faster to quantify consistency; timeouts, when present, are counted at the cap and reported separately. 

Accuracy is evaluated by comparing the \emph{optimality gap (MIPGap)} reported by the solver at termination under identical stopping criteria (same time limit and tolerances) fro the expert-written formulation and for out system-generated formulation. For the main benchmarks, we use 300 representative instances; for the small-scale benchmarks, we use 100 instances. We adopt the standard definition for minimal quality $
\mathrm{gap} \;=\; \frac{\lvert z_{\text{P}} - z_{\text{D}} \rvert}{z_{\text{P}}},$
where $z_P$ is the incumbent objective bound and $z_D$ is the best dual bound at termination. All solutions produced by our system are verified feasible with respect to the original MILP constraints, and infeasible runs (if any) are excluded from gap statistics and reported separately as validation failures.

\subsection{Results: Solver Runtime Performance}
\begin{figure}
    \centering
    \includegraphics[width=0.8\linewidth]{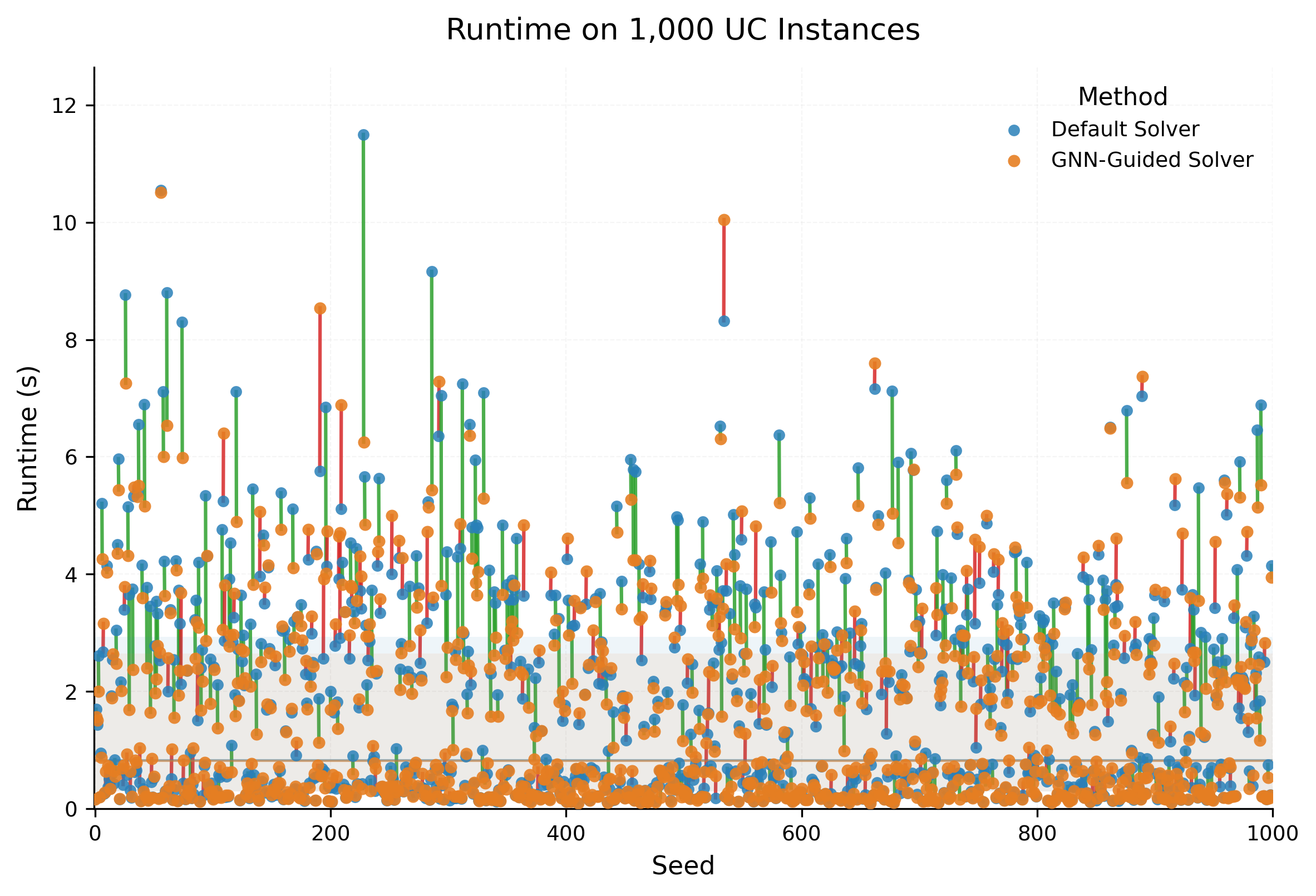}
    \caption{Comparison Results: the Improved Runtime.}
    \label{fig4}
\end{figure}

We quantify speedup using the per-instance ratio \(\tau = t_{\text{guided}}/t_{\text{default}}\) (values \(<1\) indicate improvement). On the 1{,}000-instance benchmark (Fig.~\ref{fig4}), the default solver averages \(\approx 1.79\,\mathrm{s}\) per instance, whereas our GNN guidance reduces this to \(\approx 1.64\,\mathrm{s}\), a \(1.09\times\) (\(\sim 9\%\)) gain with a similarly improved median. The histogram of \(\tau\) concentrates below \(1.0\), with a light tail above \(1.0\) driven by very easy cases where scoring overhead can offset benefits.

We observe the same pattern on 24-hour UC benchmarks spanning 3–60 generators (5{,}000 instances; Fig.~\ref{fig5:a}). Speedups persist across all size bins and become more pronounced as instance size grows, where search dominates overhead; for the smallest bins, improvements are modest and occasionally neutral. These results indicate that learned branching priorities can accelerate MILP solving in the energy domain without modifying the underlying engine. Even a \(\sim 10\%\) reduction in wall-clock time is meaningful for large-scale or time-sensitive operations, and leaves room for further gains via tuned models or adaptive refresh policies.

\subsection{Results: Solver Quality and Accuracy}

\begin{figure*}
\subfloat[Median runtime by generator count]{\label{fig5:a} \includegraphics[width=0.38\linewidth]{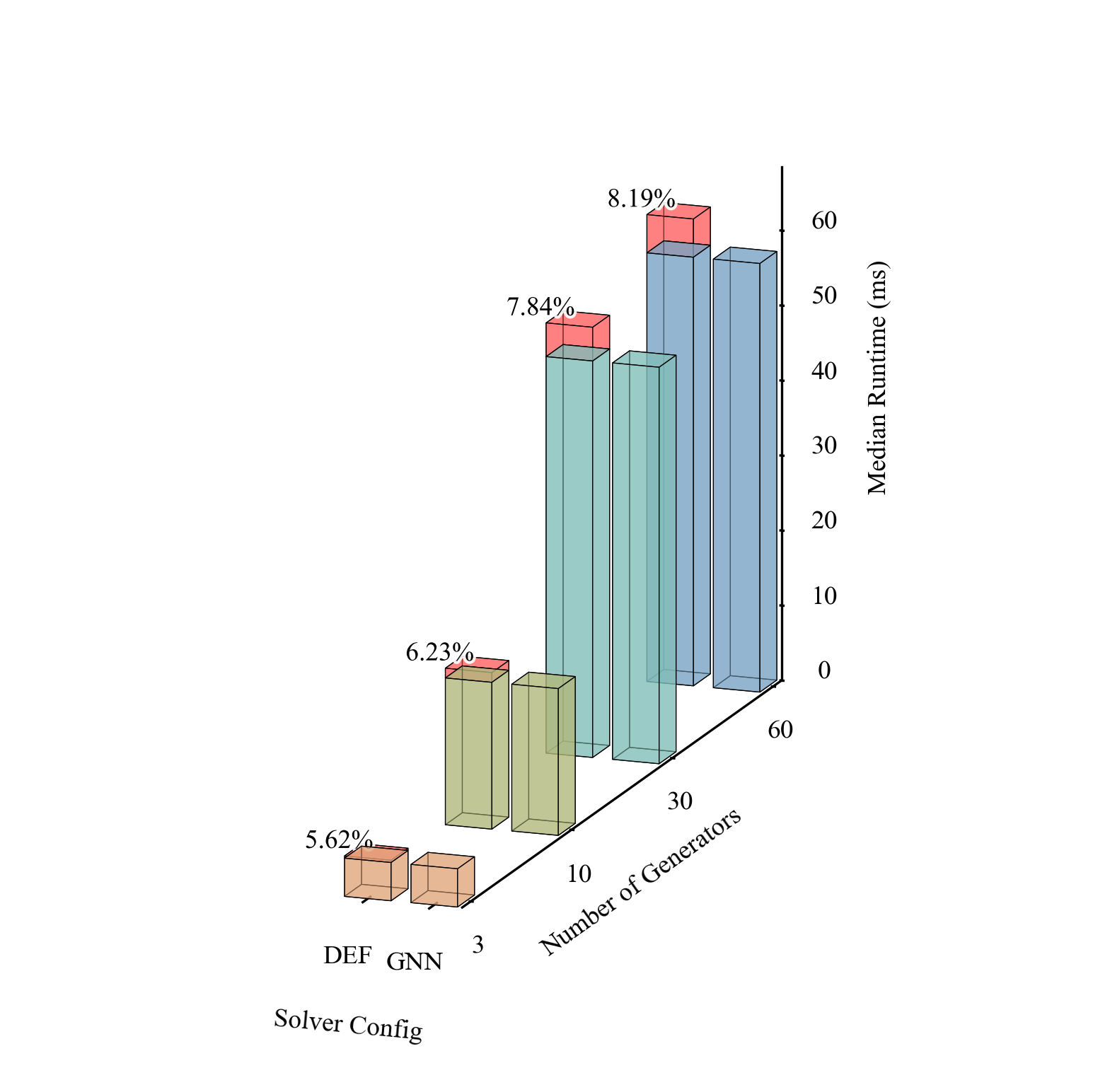}}%
\hfill
\subfloat[Optimality-gap distributions]{\label{fig5:b} \includegraphics[width=0.62\linewidth]{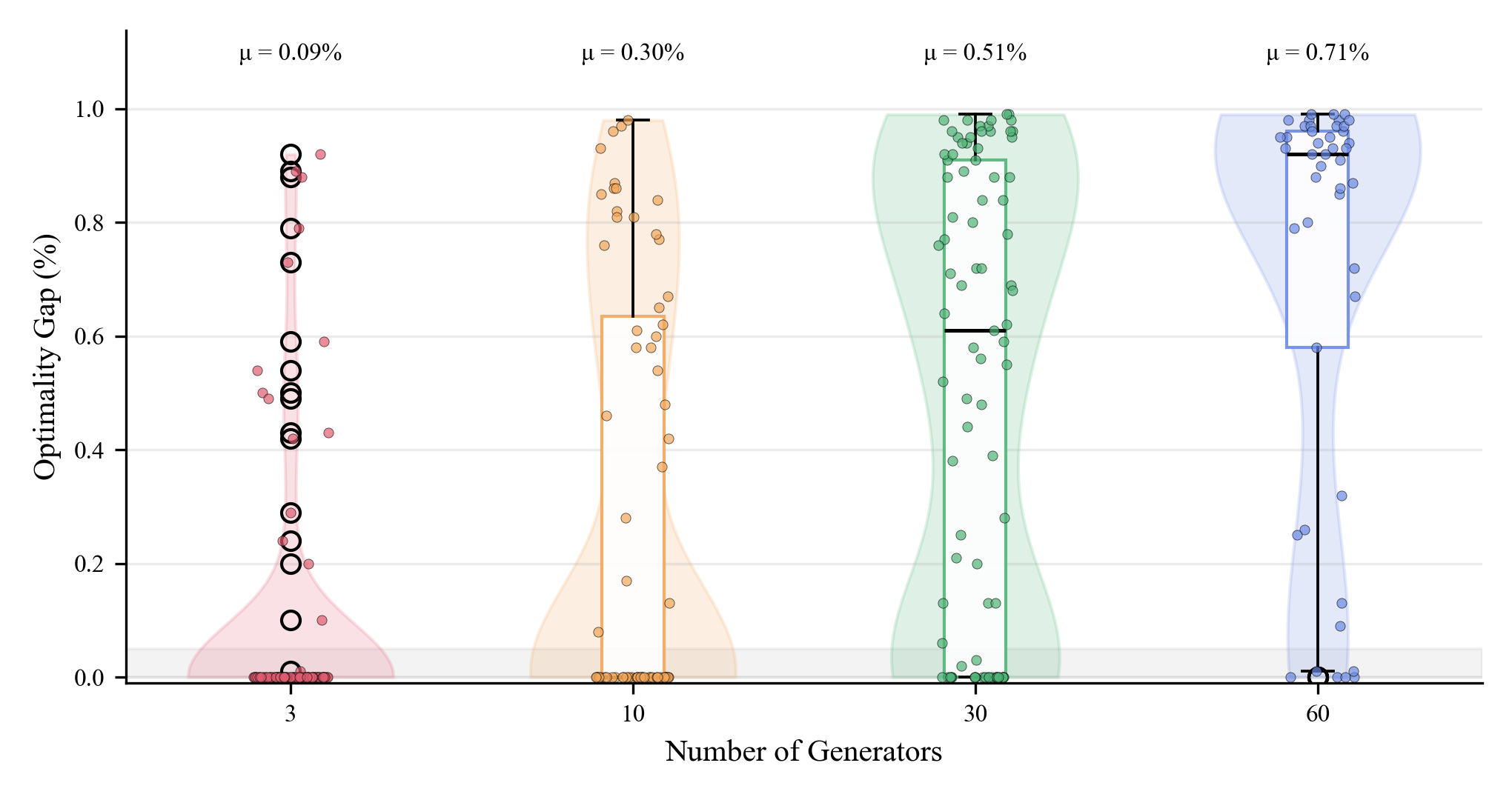}}%
\caption{Performance on Small-Scale UC Benchmarks (3/10/30/60 generators).}
\end{figure*}

We access solution quality using the solver-reported optimality gap (MIPGap) under identical tolerances and time limits.
Fig.~\ref{fig5:b} reports the small-scale set with four configurations (3, 10, 30, and 60 generators; \(\sim\)100 cases each over 24 hours). Across all four sizes, gaps cluster near zero with compact interquartile ranges (IQR) and only a few outliers. A similar pattern is observed on the 300-instance quality cohort for the 50-100 generator benchmarks, indicating consistent solution quality across scales.

\section{Conclusion}
We presented an end-to-end system that translates natural-language UC descriptions into solver-ready MILP models with schema-aware parameter synthesis, constraint templates, and iterative repair. Coupled with solver-in-the-loop validation, the approach bridges domain intent and formal optimization, achieving \(100\%\) success on our test scenarios and clearly outperforming a single-pass LLM baseline in formulation fidelity. To accelerate computation, we add GNN-guided branching that prioritizes critical decisions. Together, LLM-driven modeling and learned solver guidance are complementary—LLMs provide flexibility and domain alignment, while the MILP solver and learned policies preserve rigor and efficiency—making AI-assisted UC practical for operational settings. 

Future work will extend coverage beyond thermal UC to multi-resource scheduling (hydro, storage, and variable renewables), improve first-pass accuracy via domain-tuned prompting, and explore tighter LLM–solver coupling for per-instance configuration.

\bibliographystyle{IEEEtran}
\bibliography{reference}


\end{document}